\documentclass{article}

\usepackage{arxiv}

\usepackage[T1]{fontenc}    
\usepackage{hyperref}       
\usepackage{url}            
\usepackage{booktabs}       
\usepackage{amsfonts}       
\usepackage{nicefrac}       
\usepackage{microtype}      
\usepackage{lipsum}		
\usepackage{graphicx}
\usepackage{doi}

\usepackage{setspace}
\usepackage{indentfirst}
\usepackage{subfig}

\usepackage{amsmath}
\usepackage{mathtools}
\DeclarePairedDelimiter\norm{\lVert}{\rVert}  
\DeclareMathOperator*{\argmin}{min} 

\usepackage[noadjust]{cite}  
\usepackage{float}

\usepackage{soul}
\usepackage{color, colortbl}
\definecolor{light_aqua}{rgb}{0.75,0.87,1}

\definecolor{ttc}{rgb}{0.55, 0.0, 0.55} 

\usepackage{textcomp} 

\usepackage{algorithm}
\usepackage[noend]{algorithmic}

\usepackage{paralist} 

\usepackage{tabularx}

\newcommand{\xf}[1]{Fig.~\ref{#1}}
\newcommand{\xt}[1]{Table~\ref{#1}}
\newcommand{\xalg}[1]{Algorithm~\ref{#1}}

\newcommand{\detset}{\mathbf{\mathcal{D}}} 
\newcommand{\sset}{\mathbf{\mathcal{S}}} 
\newcommand{\apsub}{\mathbf{A}} 
\newcommand{\wt}{\mathrm{WT}}
\newcommand{\iwt}{\mathrm{iWT}} 
\newcommand{\ap}{\mathrm{AP}} 

\newcommand{\nim}{\mathbf{Y}} 
\newcommand{\im}{\hat{\mathbf{X}}} 

\newcommand{\jscale}{$J$-scale }
\newcommand{\kscale}{$K$-scale }

\title{Multiscale Sparsifying Transform Learning for Image Denoising}

\date{} 					

\usepackage{authblk}

\author[1]{Ashkan Abbasi}
\author[1]{Amirhassan Monadjemi}
\author[2]{Leyuan Fang}
\author[3]{Hossein Rabbani}
\author[1]{Neda Noormohammadi}
\author[4]{Yi Zhang}
\affil[1]{University of Isfahan, Iran}
\affil[2]{Hunan University, China}
\affil[3]{Isfahan University of Medical Sciences, Iran}
\affil[4]{Sichuan University, China}

\hypersetup{
pdftitle={Multiscale Sparsifying Transform Learning for Image Denoising},
pdfsubject={q-bio.NC, q-bio.QM},
pdfauthor={David S.~Hippocampus, Elias D.~Striatum},
pdfkeywords={First keyword, Second keyword, More},
}

\begin{document}
\maketitle

\begin{abstract}
The data-driven sparse methods such as synthesis dictionary learning (e.g., \mbox{K-SVD}) and sparsifying transform learning have been proven effective in image denoising. However, they are intrinsically single-scale which can lead to suboptimal results. 
We propose two methods developed based on wavelet subbands mixing to efficiently combine the merits of both single and multiscale methods. We show that an efficient multiscale method can be devised without the need for denoising detail subbands which substantially reduces the runtime. The proposed methods are initially derived within the framework of sparsifying transform learning denoising, and then, they are generalized to propose our multiscale extensions for the well-known \mbox{K-SVD} and SAIST image denoising methods. We analyze and assess the studied methods thoroughly and compare them with the well-known and state-of-the-art methods. 
The experiments show that our methods are able to offer good trade-offs between performance and complexity.
\end{abstract}


\section{Introduction}
Sparse models have been proven effective in various image restoration and compression problems. One approach to exploit the notion of sparsity is based on fixed transform models (such as discrete cosine transform (DCT) \cite{Ahmed_Natarajan_Rao_1974} and wavelets \cite{Mallat_1999}). In these models, the analysis and synthesis filters are derived based on a mathematical model of the data and desired characteristics, leading to highly structured transforms with fast implementations. In addition to their computational efficiency, wavelet based methods are multiscale. This can partially enriches their representation ability and widens the effective processing area.

The main steps of wavelet denoising (also known as wavelet shrinkage) \cite{Mallat_1999} are transforming the noisy image, denoising detail subbands through a thresholding operator, and computing the inverse wavelet transform. Despite the advantages of these methods, annoying visual artifacts are inevitable, which can be attributable to 1) rough frequency cut-offs due to thresholding operations, and 2) restricted representation ability. Statistical modeling is the main approach to improve the quality of thresholding \cite{Chang2000, Donoho_Johnstone_1995, Donoho_Johnstone_1994}. It has been shown that considering the intra-scale and inter-scale correlations of wavelet coefficients is very effective (e.g., BLS-GSM \cite{Portilla_Strela_Wainwright_Simoncelli_2003}). There are also attempts to reduce artifacts through variational methods \cite{Durand_Froment_2001, Li_Ghosal_2015}, introducing redundancy in the denoising process \cite{Coifman_Donoho_1995}, and using redundant wavelet transforms \cite{Fowler_2005,Selesnick_Baraniuk_Kingsbury_2005,Starck_Fadili_Murtagh_2007,Starck_Murtagh_Fadili_2015}.

Despite tremendous efforts made over the years to improve wavelet denoising, their representation ability is intrinsically restricted since these transforms are not data adaptive. In contrast, synthesis sparse models \cite{Aharon_Elad_Bruckstein_2006,Elad_Aharon_2006,Olshausen_Field_1996}) provide an adaptive framework to exploit the notion of sparsity. 
To enjoy both simplicity of sparse coding and adaptivity to data, the authors in \cite{Ravishankar_Bresler_2013a, Ravishankar_Bresler_2013b, Ravishankar_Bresler_2015} present the sparsifying transform learning model. They have shown that their model has comparable or even better performance compared to the well-known \mbox{K-SVD} synthesis dictionary learning \cite{Aharon_Elad_Bruckstein_2006,Elad_Aharon_2006}. However, both mentioned data-driven sparse models (dictionary learning and sparsifying transform learning) are inherently single-scale.

Image denoising methods often involve patch-based (or local) operations. When noise is weak, local modeling can achieve plausible results. However, as noise level increases, local image structures are substantially distorted by noise, and thus denoising through local operations becomes difficult. In this case, enlarging the effective modeling and denoising regions through multiscale processing is effective \cite{Burger_Harmeling_2011,Facciolo_Pierazzo_Morel_2017,Feng_Qiao_Xi_Chen_2016,Lebrun_Colom_Buades_Morel_2012,Lebrun_Colom_Morel_2015,Lefkimmiatis_Maragos_Papandreou_2009,Mairal_Sapiro_Elad_2008,Ophir_Lustig_Elad_2011,Papyan_Elad_2016,Pierazzo_Morel_Facciolo_2017,Rajashekar_Simoncelli_2009,Remez_Litany_Giryes_Bronstein_2018,Sulam_Ophir_Elad_2014,Xianming2014,Zontak_Mosseri_Irani_2013}. Also, it is shown that focusing on removing noise from high-frequency contents in the most patch-based methods lead to considerable artifacts and low-frequency content loss \cite{Burger_Harmeling_2011,Facciolo_Pierazzo_Morel_2017,Lebrun_Colom_Buades_Morel_2012,Lebrun_Colom_Morel_2015,Papyan_Elad_2016,Pierazzo_Morel_Facciolo_2017}. Even non-local methods such as non-local means \cite{Buades_Coll_Morel_2005a,Buades_Coll_Morel_2005b,Buades_Coll_Morel_2011} and BM3D \cite{Dabov_Foi_Katkovnik_Egiazarian_2007} could not sufficiently reduce these types of artifacts \cite{Papyan_Elad_2016,Sulam_Ophir_Elad_2014}. Nevertheless, promising results have been reported in several multiscale methods \cite{Burger_Harmeling_2011,Facciolo_Pierazzo_Morel_2017,Mairal_Sapiro_Elad_2008,Ophir_Lustig_Elad_2011,Papyan_Elad_2016,Pierazzo_Morel_Facciolo_2017,Sulam_Ophir_Elad_2014}. Some representative works include the multiscale meta-procedure proposed by \cite{Burger_Harmeling_2011}, multiscale \mbox{K-SVD} (\mbox{MS K-SVD}) \cite{Ophir_Lustig_Elad_2011}, \mbox{Fused K-SVD} \cite{Sulam_Ophir_Elad_2014}, multiscale EPLL (MSEPLL) \cite{Papyan_Elad_2016}, conservative scale recomposition \cite{Facciolo_Pierazzo_Morel_2017}, MS DCT \cite{Pierazzo_Morel_Facciolo_2017}, and MSND \cite{Feng_Qiao_Xi_Chen_2016}. Most of these methods adopt the pyramid image representation as a way to construct their multiscale methods. In contrast, \mbox{MS K-SVD} and \mbox{Fused K-SVD} exploit the wavelet transform. Since various forms of wavelet transforms and filter banks were proposed over the years, these approaches have a potential to boost denoising methods for different types of images. However, the former generally suffers from considerable artifacts, and the latter is computationally demanding. 

Here, our goal is to analyze and extend the approach of subbands denoising which was initially proposed by \mbox{MS K-SVD} \cite{Ophir_Lustig_Elad_2011}. In our first proposed method, a fusion method based on the wavelet subbands mixing  \cite{Coupé_Hellier_Prima_Kervrann_Barillot_2008,Lukin_2006} is developed to combine the results of single and multiscale methods. 
Although this method is simple, it greatly suppresses undesirable artifacts and maintains the low-frequency and main structures of the image. Additionally, unlike the joint sparse representations model proposed by \mbox{Fused K-SVD} \cite{Sulam_Ophir_Elad_2014}, it is very fast with favorably comparable results. Nevertheless, the whole method is still costly since it needs to denoise the image itself and all of its subbands. Therefore, we simplify the operations and develop a new method in which there is no need to explicitly denoise detail subbands. 
This latter method mainly saves the computational cost of denoising detail subbands, and it shows a systematic way of exploiting the low-pass subbands of discrete wavelet transform for image denoising. 

We initially introduce our multiscale methods in the context of sparsifying transform learning denoising (TLD) \cite{Ravishankar_Bresler_2013a, Ravishankar_Bresler_2013b, Ravishankar_Bresler_2015} due to its efficiency and numerical stability. 
Then, we generalize them to propose mixing based multiscale methods based on \mbox{K-SVD} and spatially adaptive iterative singular-value thresholding (SAIST) \cite{Dong_Shi_Li_2013}. 
We analyze and examine our studied methods comprehensively.
On the task of gray-scale image denoising, our multiscale SAIST can compete well with the state-of-the-art methods such as SSC-GSM \cite{Dong_Shi_Ma_Li_2015}, STROLLR \cite{strollr_tip2020}, and \mbox{GSRC-NLP} \cite{Zha_Yuan_Wen_Zhou_Zhu_2020}. The overall qualitative and quantitative comparisons reveal that our methods offer a good trade-off between performance and complexity.

The rest of this paper is organized as follows. In the following section, we briefly review the related works. Next, we describe the studied multiscale methods in Section \ref{methods}. Then, the experimental results are presented in Section \ref{exp} and the paper is concluded in Section \ref{conc}.

\section{Related Works}\label{related}
In this section, 
we briefly review some previous works on the multiscale data-driven sparsity based methods. 
Next, we take a look at the framework of sparsifying transform learning denoising (TLD) \cite{Ravishankar_Bresler_2013a, Ravishankar_Bresler_2013b, Ravishankar_Bresler_2015}, since it is used as our main single-scale baseline to introduce our multiscale extensions.

\subsection{Multiscale Data-driven Sparse Models}\label{related:sparse}
There are promising efforts to integrate multiscale analysis with data-driven sparse models \cite{Bacchelli_Papi_2006,Fang_Li_Nie_Izatt_Toth_Farsiu_2012,Hughes_Rockmore_Wang_2013,Mairal_Sapiro_Elad_2008,Olshausen_Sallee_Lewicki_2001,Ophir_Lustig_Elad_2011,Sallee_Olshausen_2003,Sulam_Ophir_Elad_2014,Yan_Shao_Liu_2013}. The idea of learning sparse multiscale image representations dates back to the works of \cite{Olshausen_Sallee_Lewicki_2001,Sallee_Olshausen_2003}. They show how a wavelet basis can be adapted to the statistics of natural images, and achieve slightly better coding efficiency \cite{Olshausen_Sallee_Lewicki_2001} and denoising \cite{Sallee_Olshausen_2003}. But these methods are fairly elaborate and computationally demanding since they rely on the sampling techniques to infer sparse codes. The work by \cite{Mairal_Sapiro_Elad_2008} extends \mbox{K-SVD} \cite{Aharon_Elad_Bruckstein_2006,Elad_Aharon_2006} to simultaneously use different sizes of atoms, each size corresponding to a different scale. Their method achieves very promising results, though it is computationally expensive.

In a different approach, data-driven sparse models are directly applied in the multi-resolution analysis domain \cite{Bacchelli_Papi_2006,Fang_Li_Nie_Izatt_Toth_Farsiu_2012,Hughes_Rockmore_Wang_2013,Mairal_Sapiro_Elad_2008,Ophir_Lustig_Elad_2011,Sulam_Ophir_Elad_2014,Yan_Shao_Liu_2013}.  In this way, the denoising method acts as a complex wavelet shrinkage operator and the frequency selectivity of wavelet subbands acts as a divide and conquer strategy which could result in sparser representations. In \cite{Bacchelli_Papi_2006}, a method based on filtering principal component coefficients of wavelet packet subbands is presented. The work in \cite{Ophir_Lustig_Elad_2011} has successfully shown that the \mbox{K-SVD} denoising method \cite{Elad_Aharon_2006} can be directly used to filter wavelet subbands. Their experiments show that \mbox{MS K-SVD} partially outperforms single-scale KSVD \cite{Elad_Aharon_2006} due to better recovery of textures, edges, and main image structures. However, \mbox{MS K-SVD} \textquotesingle s results suffer from artifacts \cite{Hughes_Rockmore_Wang_2013,Sulam_Ophir_Elad_2014,Yan_Shao_Liu_2013}. The same approach is used in \cite{Fang_Li_Nie_Izatt_Toth_Farsiu_2012,Hughes_Rockmore_Wang_2013,Yan_Shao_Liu_2013} by considering other techniques such as structural clustering, nonlocal self-similarity, and Bayesian learning of a dictionary in the domain of the Laplacian pyramid \cite{Burt_Adelson_1983}.

Instead of trying to enrich the subband denoising approach with different techniques, \mbox{Fused K-SVD} \cite{Sulam_Ophir_Elad_2014} uses a joint sparse representation based method to fuse the results of the single \cite{Elad_Aharon_2006} and multiscale \mbox{K-SVD} \cite{Ophir_Lustig_Elad_2011} methods. The assumption is that the underlying structure between these two results are the same. Therefore, a joint sparse representation technique \cite{Yu_Qiu_Ren_2013} can be used to recover the underlying image without artifacts. Although the artifacts are suppressed greatly, the computational cost increases significantly since the joint sparse representations are obtained using concatenated input vectors and dictionaries. We will show that this step can be efficiently replaced by a method based on the subbands mixing technique. As we will demonstrate in our experiments, the proposed method has favorably comparable performance in comparison with \mbox{Fused K-SVD}, while its computation is greatly cheaper.

\subsection{Sparsifying Transform Learning Denoising (TLD)}\label{related:tld}
Transform models \cite{Ravishankar_Bresler_2013a, Ravishankar_Bresler_2013b, Ravishankar_Bresler_2015} assumes that signals of interest can be approximately sparsifiable using a transform matrix. Given a square transform matrix W and sparse code x, the least squares estimate of the underlying signal vector u is obtained using $u = W^\dagger x$, where $W^\dagger$ denotes pseudoinverse of $W$. Ideally, in the transform model, corruptions such as noise cannot be well approximated in the transform domain, thus finding the sparse code of a corrupted signal over a suitable transform helps to restore it. Therefore, in \cite{Ravishankar_Bresler_2013b}, an adaptive signal denoising method is proposed as follows:

\begin{equation}\label{eq:tld}
	\begin{aligned}
		& \argmin_{W,X,U} \norm{WY - X}^2_F + \lambda Q(W) + \norm{Y - U}\\
		& \mathrm{subject\;to\quad} \norm{X_i}_0 \le l \quad \forall i
	\end{aligned}
\end{equation}
where $Y \in \mathbb{R}^{n \times N}$ is a matrix whose columns represent $N$ signals corrupted by an additive white Gaussian noise, and $\hat{U} \in \mathbb{R}^{n \times N}$ is a matrix containing recovered signals, $Q(W) = -\log\lvert \mathrm{det\,}W \rvert + \mu\norm{W}_F^2$ is a regularizer to prevent trivial solutions, and the parameter $\tau$ is chosen inversely proportional to noise level $\sigma$.

To use the above formula for image denoising, the observed image $\nim$ is first divided using $R_iy$ into small overlapping patches of size $\sqrt{n} \times \sqrt{n}$, with $R_i$ being the patch extraction operator and y being the vector form of image. Then, the mean intensity of each patch is removed, and all extracted patches $Y_i = R_iy$ are arranged as columns of the matrix $Y$. Next, the problem (\ref{eq:tld}) is solved using an alternating optimization algorithm which alternates between a transform learning step and a variable sparsity update step.

The transform learning step is composed of a sparse coding step and a transform update step. Both steps are exactly solved using closed-form solutions \cite{Ravishankar_Bresler_2015}. The sparse codes $X_i$ are obtained by thresholding $WY_i$ to the $l_i$ largest magnitude coefficients, and the transform update can be solved through a closed-form solution with global convergence guarantee.

In practice, it turns out that using a fixed sparsity level degrades the denoising performance. Therefore, an efficient variable sparsity update step is also proposed in \cite{Ravishankar_Bresler_2013a, Ravishankar_Bresler_2013b}. For fixed $W$ and $X_i\;(i=1,\dots,N)$, this step reduces to the least squares problem in the $\hat{U}_i$\textquotesingle s. Each $\hat{U}_i$ is updated by adding one nonzero element at a time from $WY_i$ to $X_i$, until the error measure $\norm{Y_i - \hat{U}_i}$ falls below a given threshold $nc^2\sigma^2$; where $c$ is a fixed constant which controls the sparsity of representations. Finally, when all the estimated patches $\hat{U}_i$ are recovered, the denoised image $\hat{\mathbf{U}}$ can be obtained by adding back the mean intensities to the final estimates and averaging them at their respective locations in the image.

For discussing about TLD\textquotesingle s initialization, stopping criteria, and convergence guarantee, we refer the interested reader to its mentioned original papers (e.g., \cite{Ravishankar_Bresler_2013a}).

\section{The Methods}\label{methods}
\subsection{Multiscale TLD (MTLD)}\label{methods:mtld}
Let\textquotesingle s create a baseline multiscale denoiser by exploiting the idea of \mbox{MS K-SVD} \cite{Ophir_Lustig_Elad_2011}. However, to make running experiments faster, we use TLD in place of \mbox{K-SVD} and develop multiscale TLD (MTLD). 

To derive MTLD, we first begin by applying an analytical transform $\Phi$ to the data matrix Y in the formula of sparsifying transform learning (\ref{eq:tld}). Let the transform matrix $W$ in (\ref{eq:tld}) be the product of two different square transforms $B\Phi$, then:

\begin{equation}\label{eq:tldb}
	\begin{aligned}
		& \argmin_{B,X,U} \norm{B\Phi Y - X}^2_F + \lambda Q(B) + \norm{\Phi Y - U}\\
		& \mathrm{subject\;to\quad} \norm{X_i}_0 \le l \quad \forall i
	\end{aligned}
\end{equation}
where, $\Phi$ is the fixed analytical transform (e.g., DCT or wavelet), and B is a transform matrix that is to be learned from transformed data. If $\Phi$ is a wavelet transform, we can rewrite the problem for each subband separately:

\begin{equation}\label{eq:tldc}
	\begin{aligned}
		& \argmin_{B_s,X_s,\hat{U}_s} \norm{B_s \Phi_s U_s  -  X_s}^2_F + \lambda Q(B_s) + \norm{\Phi_s Y - U_s}\\
		& \mathrm{subject\;to\quad} \norm{X_{s,i}}_0 \le l \quad \forall i, \forall s
	\end{aligned}
\end{equation}
where subscript $s$ indicates different wavelet subbands, $\Phi_s Y$ contains vectors extracted form the $s$-th wavelet subband, and $U_s$ contains the recovered vectors.

\subsubsection{MTLD Steps and Its Parameters}\label{methods_mtld_steps}


We can realize the idea of denoising subbands (Eq. \ref{eq:tldc}) using \xalg{alg_mtld}. At the first line, it applies a \jscale wavelet transform (denoted by $\wt(.)$) to the input image $\nim$ and decomposes it into a set of subbands $\sset = \{\apsub^J, \detset^J\}$ which includes the approximate subband at scale $J$ ($\apsub^J$) and a set ($\detset^J$) containing the detail subbands. Next, each subband is denoised via TLD in line \ref{alg_mtld_denoise_subbands}. Then, the denoised image ($\im_M$) is reconstructed by computing the inverse wavelet transform ($\iwt(.)$) at line \ref{alg_mtld_output}.

\begin{algorithm}
	\caption{
		multiscale TLD (MTLD)
	}
	\label{alg_mtld}
	\begin{algorithmic}[1]
		\REQUIRE Noisy image ($\nim$), Noise levels of subbands ($\sigma_\mathbf{S}$), Patch size ($p$), Number of scales ($J$).
		\ENSURE Type of wavelet transform ($\wt$) and its filterbank ($af$ and $sf$).
		\STATE $\sset = \wt(\nim,J,af)$ \hfil \COMMENT{$\sset$ will be  $\{\apsub^J,\detset^J\}$}
		\STATE $\hat{\sset} = \{\}$\\
		\FORALL{subband ($\mathbf{S} \in \sset$)}
		\STATE $\hat{\mathbf{S}} = \mathrm{TLD}(\mathbf{S},\sigma_\mathbf{S},p)$ \label{alg_mtld_denoise_subbands}
		\STATE $\hat{\sset} = \hat{\sset} \cup \hat{\mathbf{S}}$
		\ENDFOR
		\RETURN $\im_M = \iwt(\hat{\sset},J,sf)$ \label{alg_mtld_output}
	\end{algorithmic}
\end{algorithm}

For implementing this method in practice, it is required to determine the type of the \jscale wavelet transform and its filter bank. These are hyper-parameters. While various wavelet transforms can be used to realize MTLD, here, $\wt(.)$ function is substituted with a discrete wavelet transform (DWT) \cite{Mallat_1999}, and the discrete Meyer analysis and synthesis filters are used.

Patch size ($p$) is usually set to 11 x 11 pixels. The noise level ($\sigma$) is commonly assumed to be known or can be estimated by a noise estimation method \cite{Foi_2011,Rakhshanfar_Amer_2016}. We experimentally found that we can use the noise level of the input image for denoising its subbands $\sigma_\mathbf{S} = \sigma$.

\subsubsection{Motivational Example: MTLD versus TLD}\label{methods_mtld_motiv}

\begin{figure}[h!]
	\centering
	\includegraphics[height = 9cm]{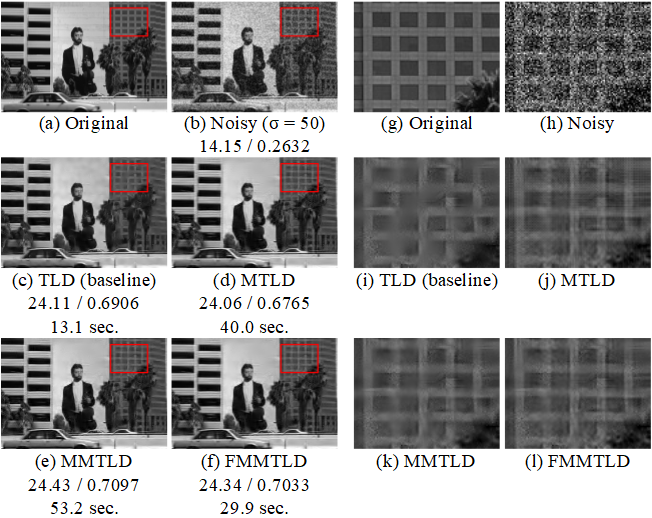}
	\caption{Visual comparison of the denoising results by the studied methods. For each method, its PSNR, SSIM, and runtime are reported below it}
	\label{fig_motiv}
\end{figure}

We show the outputs of TLD and MTLD for denoising a test image in Fig. \ref{fig_motiv} (b). In this example, MTLD is implemented with a 1-scale ($J$=1) DWT, and the results are shown in Fig. \ref{fig_motiv} (c) and (d). These results confirm that although the main image structures (or low-frequency information) are better recovered through MTLD, the artifacts in MTLD\textquotesingle s output negatively affect the result. The quality of denoised images are quantized using the Peak Signal-to-Noise-Ratio (PSNR) and structural similarity index (SSIM) \cite{Wang_Bovik_Sheikh_Simoncelli_2004}.

A number of reasons can be thought of for explaining why MTLD generally leads to inferior results than TLD: First, denoising a subband is not an easy task due to its low signal to noise ratio \cite{Papyan_Elad_2016}. This is why a lot of efforts have been made over the years to introduce more effective wavelet shrinkage methods. Second, to denoise subbands using TLD, careful parameter adjustment for each subband might be required, which is a quite cumbersome procedure. These reasons may negatively contribute to the overall quality of MTLD.


\begin{figure*}
	\centering
	\includegraphics[height = 5.5cm]{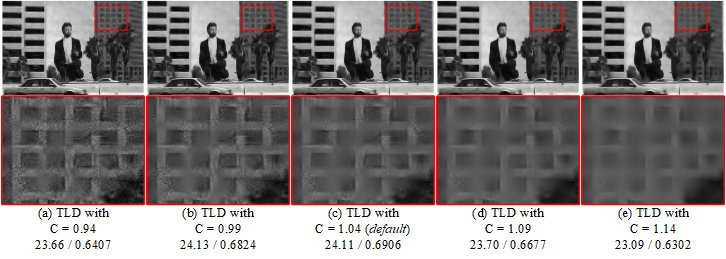}
	\caption{Effects of using different sparsity levels (or error thresholds) on the output of TLD. The $c$ constant (Section \ref{related:tld}) modifies the sparsity level in TLD. By increasing this constant, the representations in TLD becomes sparser, and, therefore, noise suppression becomes stronger. For each method, its PSNR and SSIM are reported below it.}
	\label{fig_const}
\end{figure*}

In order to restore the damaged structures in Fig. \ref{fig_motiv} (c), a reader may suggest carefully controlling the amount of denoising in TLD. This can be done by changing the $c$ constant in TLD (Section \ref{related:tld}). This constant controls the sparsity level, and it has a direct impact on the amount of denoising. To visually inspect the effects of different values of this hyper-parameter, we change it with steps of 0.05 in Fig. \ref{fig_const}. Comparing Fig. \ref{fig_const} (b) and (c), it can be seen that although the default value of $c$ is reasonable for this specific image, when we set $c$ to 0.99 (Fig. \ref{fig_const} (b)), PSNR maximizes, and the damaged structures are better recovered. However, note that this is achieved with the expense of decreasing SSIM and tolerating more artifacts. It turns out that even with this careful hyper-parameter adjustment, TLD cannot efficiently recover the damaged parts without sacrificing the quality.

\subsection{
	Proposed Mixed MTLD (MMTLD) 
}\label{methods:mmtld}
Using wavelet subbands mixing technique \cite{Coupé_Hellier_Prima_Kervrann_Barillot_2008,Lukin_2006}, we can combine the merits of both TLD and MTLD: (i) The main image structure and low frequency contents of MTLD\textquotesingle s output, and (ii) The high frequency contents of TLD\textquotesingle s output. We call this method the Mixed MTLD (MMTLD) method.

The main steps of MMTLD is listed in \xalg{alg_mmtld_1}. At first, the input image is denoised through both TLD and MTLD (lines \ref{alg_mmtld_1_tld} and \ref{alg_mmtld_1_mtld}). Next, their outputs are separately decomposed using a \kscale wavelet transform (denoted by $\wt_X(.)$). Then, the wavelet subbands mixing is applied (in line \ref{alg_mmtld_1_mix}) to create a new set ($\hat{\sset}_X$) which comprises of the approximate subband ($\hat{\apsub}_M^K$) of MTLD and detail subbands ($\hat{\detset}_S^K$) of TLD. Finally, the inverse wavelet transform is computed on ($\hat{\sset}_X$) to reconstruct the final denoised image ($\im_{MM}$).

\begin{algorithm}[H]
	\caption{
		The first version of mixed MTLD (MMTLD) denoted by MMTLD-1
	}
	\label{alg_mmtld_1}
	\begin{algorithmic}[1]
		\REQUIRE Noisy image ($\nim$) and its noise level ($\sigma$), Patch size ($p$), Number of scales ($J$), Number of scales ($K$) for the mixing stage.
		\ENSURE Hyper-parameters of \xalg{alg_mtld}, Type of wavelet transform ($\wt_X$) and its filterbank ($af$ and $sf$).
		\STATE $\im_S = \mathrm{TLD}(\nim,\sigma,p)$ \label{alg_mmtld_1_tld}
		\STATE $\im_M = \mathrm{MTLD}(\nim,\sigma,p,J)$ \label{alg_mmtld_1_mtld}
		\STATE $\hat{\sset}_S = \wt_X(\hat{\mathbf{X}}_S,K,af)$ \hfil \COMMENT{$\hat{\sset}_S$ will be  $\{\hat{\apsub}^K_S,\hat{\detset}^K_S\}$} \label{chp5_alg_mmtld_1_WT_X_s}
		\STATE $\hat{\detset}_S^K = $ Get the detail subbands from $\hat{\sset}_S$
		\STATE $\hat{\sset}_M = \wt_X(\im_M,K,af)$ \hfil \COMMENT{$\hat{\sset}_M$ will be  $\{\hat{\apsub}^K_M,\hat{\detset}^K_M\}$} \label{alg_mmtld_1_WT_X_m}
		\STATE $\hat{\apsub}_M^K = $ Get the approximate subband from $\hat{\sset}_M $ \label{chp5_alg_mmtld_1_getA_M}
		\STATE $\hat{S}_X = \{ \hat{\apsub}_M^K, \hat{\detset}_S^K \}$ \label{alg_mmtld_1_mix}
		\RETURN $\im_{MM} = \iwt_X(\hat{\sset}_X,K,sf)$
	\end{algorithmic}
\end{algorithm}

To demonstrate how the mixing stage of MMTLD-1 works in practice, let\textquotesingle s implement the \kscale transform function $\wt_X(.)$ used for mixing subbands with a 1-scale ($K$=1) isotropic undecimated wavelet transform (IUWT) with an Astro filter bank \cite{Starck_Fadili_Murtagh_2007}. In each level, IUWT decomposes the input into one approximate and one detail subbands. \xf{fig_motiv} (c) and (d) show the outputs of TLD and MTLD, respectively. These outputs are decomposed by a $1$-scale IUWT in \xf{fig:subbands}. In this figure, the first two images show the subbands obtained from TLD\textquotesingle s output (denoted by $\hat{\sset}_S$ in \xalg{alg_mmtld_1}) and the second two images show the subbands of MTLD\textquotesingle s output ($\hat{\sset}_M$). This figure clearly show that the low frequency contents in the approximate subband of MTLD (c) is much richer than the corresponding subband of TLD (a). Conversely, the detail subband of MTLD (d) is noisier than corresponding subband of TLD (b). Therefore, line \ref{alg_mmtld_1_mix} of MMTLD-1 (\xalg{alg_mmtld_1}) takes the strength of TLD and MTLD (\xf{fig:subbands} (b) and (c)) and discards their weaknesses ((a) and (d)) to produce the final output (\xf{fig_motiv} (e)).

\begin{figure}[h!]
	\centering
	\includegraphics[width=0.8\textwidth]{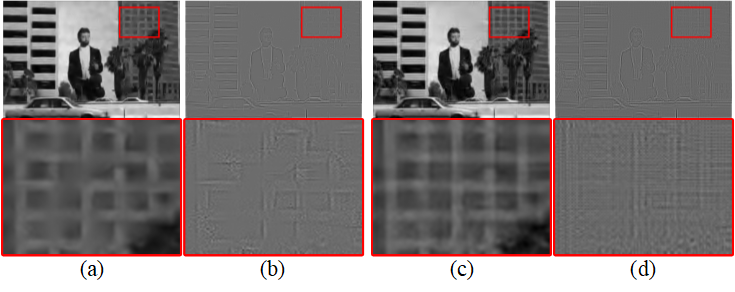}
	\caption{Illustration of the subbands obtained by applying 1-scale ($K$=1) IUWT decompositions on TLD and MTLD outputs\textquotesingle. In the first two images, the TLD subbands are shown: (a) the approximate subband ($\hat{\apsub}_S^K$), and (b) the detail subband ($\hat{\detset}_S^K$). In the last two images, the subbands of MTLD \textquotesingle s output are shown: (c) the approximate subband ($\hat{\apsub}_M^K$), and (d) the detail subband ($\hat{\detset}_M^K$).}
	\label{fig:subbands}
\end{figure}

\subsection{Proposed Fast MMTLD (FMMTLD)}\label{methods:fmmtld}
In MMTLD-1 (\xalg{alg_mmtld_1}), there are two types of wavelet transforms: 1- $\wt(.)$ is a \jscale transform used in the denoising stage of MTLD, and 2- $\wt_X(.)$ is a \kscale transform used in the mixing stage. These two transforms are completely independent of each other, and we are free to choose the appropriate transform for each stage. However, when these two transforms are selected to be the same, considerable amount of computations become redundant.

In MMTLD-1 subbands of MTLD are extracted in line \ref{alg_mmtld_1_WT_X_m}. When both wavelet transforms ($\wt(.)$ and $\wt_X(.)$) are the same, the $\hat{\sset}$ in MTLD and $\hat{\sset_M}$ in MMTLD-1 becomes equal. Therefore, the inverse wavelet transform at the end of MTLD becomes redundant, and we can use $\hat{\sset_M} = \hat{\sset}$ in MMTLD-1. Additionally, in MMTLD-1 only the approximate subband from $\hat{\sset}$ is used. This implies that there is no need to denoise the set of detail subbands ($\hat{\mathbf{\detset}}_M^K$) in MTLD.

Using the same wavelet transforms ($\wt(.)$ and $\wt_X(.)$), we can remove the extra computations in MMTLD-1, and propose a more computationally efficient method. We call it fast MMTLD (FMMTLD). The steps of our first version of FMMTLD is listed in \xalg{alg_fmmtld_1}. 

In the first three lines of \xalg{alg_fmmtld_1}, the input image ($\nim$) is denoised using TLD and then its 
detail subbands ($\hat{\detset}_S^J$) are extracted. Therefore, the detail subbands are denoised indirectly. Next, in the next three lines, the approximate subband is extracted and directly denoised through TLD. Then, the method forms a new set ($\hat{\sset}_X$) from the denoised subbands. Finally, the inverse wavelet transform is computed to reconstruct the denoised image ($\im_{FMM}$). 

\begin{algorithm}
	\caption{The first version of the Fast MMTLD method denoted by FMMTLD-1}
	\label{alg_fmmtld_1}
	\begin{algorithmic}[1]
		\REQUIRE Noisy image ($\nim$) and its noise level ($\sigma$), Patch size ($p$), Number of scales ($J$).
		\ENSURE Type of wavelet transform ($\wt$) and its filterbank ($af$ and $sf$).
		\STATE $\im_S = \mathrm{TLD}(\nim,\sigma,p)$
		\STATE $\hat{\sset_S} = \wt(\im_S,J,af)$ \hfil \COMMENT{$\hat{\sset}_S$ will be  $\{\hat{\apsub}^J_S,\hat{\detset}^J_S\}$}
		\STATE $\hat{\detset}_S^J =$ Get the detail subbands from $\hat{\sset}_S$
		\STATE $\sset = \wt(\nim,J,af)$ \hfil \COMMENT{$\sset$ will be $\{ \apsub^J,\detset^J \}$}
		\STATE $\apsub^J = $ Get the approximate subband form $\sset$
		\STATE $\hat{\apsub}^J = \mathrm{TLD}(\apsub^J,\sigma,p)$\\
		\STATE $\hat{\sset}_X = \{ \hat{\apsub}^J. \hat{\detset}^J_S  \}$ \label{alg_fmmtld_1_sset_X}
		\RETURN $\im_{FMM} = \iwt(\hat{\sset}_X,J,sf)$ \label{alg_fmmtld_1_iwt}
	\end{algorithmic}
\end{algorithm}

A representative output of FMMTLD-1 (implemented with a $1$-scale DWT) is shown in \xf{fig_motiv} (f). It shows that FMMTLD-1 performs better than both TLD and MTLD, but worse than MMTLD-1. Better performance of MMTLD-1 can be attributable to the utilization of IUWT in its mixing stage. IUWT preserves translation-invariance property \cite{Starck_Fadili_Murtagh_2007}, and therefore it has less artifacts than DWT.

\subsection{Increasing the Number of Scales - Complete Versions of the Proposed Methods}\label{methods:inc}

In our motivational example, all of the results (shown in \xf{fig_motiv}) are obtained using 1-scale transforms. Naturally, we expect that by increasing the number of scales, more details can be recovered. However, in practice, the opposite happens for our proposed MMTLD-1 and FMMTLD-1 methods. In \xf{fig:increase} (e) and (f), it can be seen that using 4-scale transforms significantly reduce the effectiveness of the proposed methods. These two methods directly denoise the approximate subband and extract the denoised detail subbands from the TLD's output. By increasing the number of scales, a smaller range of low frequency information is included in the approximate subband, and the contribution of the detail subbands increases. Therefore, the quality of our proposed methods tend to TLD.

\begin{figure*}[!h]
	\centering
	\includegraphics[height = 4.3cm]{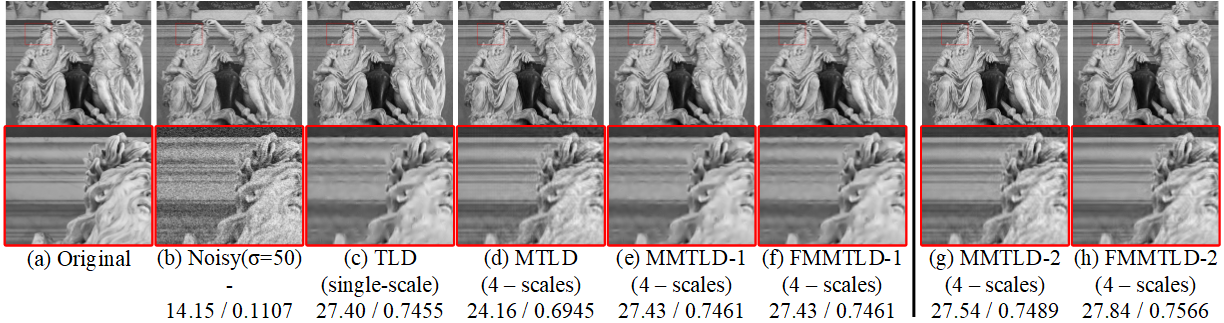}
	\caption{Effect of increasing the number of scales on the performance of the proposed multiscale methods. Setting the number of scales to 4 (J=K=4) significantly reduces the effectiveness of the multiscale processing in MMTLD-1 (e) and FMMTLD-1 (f). This weakness is solved in the second versions of these methods. The results of MMTLD-2 and FMMTLD-2 are shown in (g) and (h), respectively. The original image (a) is a high-resolution image (with spatial size of 2040x1740 pixels) from the validation set of DIV2K dataset \cite{Agustsson_Timofte_2017,Ignatov2018}. The image (a) is corrupted in (b) by adding a white Gaussian noise with $\sigma$ = 50. For each method, its PSNR and SSIM are reported below it.}
	\label{fig:increase}
\end{figure*}


To fix FMMTLD-1, the detail subbands should be processed separately, similar to MTLD. However, in contrast to MTLD, the processing should be done indirectly. Consider a wavelet transform with two scales ($J$ = 2). Then, the subbands are $\{\apsub^2,\detset^2\}$. Let\textquotesingle s show the set of detail subbands of the first level separately: $\{\apsub^2,\detset^2,\detset^1 \}$. Given this set, the inverse transform is computed as depicted in the right side of \xf{fig_schematic} (a) \cite{Mallat_1989}. The mixing stage of FMMTLD-1 (\xalg{alg_fmmtld_1}) is essentially the inverse wavelet transform with selected denoised subbands (lines \ref{alg_fmmtld_1_sset_X} and \ref{alg_fmmtld_1_iwt}). Now, instead of using the last approximate subband and detail subbands, let\textquotesingle s assume that we store the input image ($\apsub^0 = \nim$) as the zero-th approximate subband and all of its approximate subbands across different scales: $\{\apsub^1,\apsub^2\}$ . In this way, no information is lost. Then, to perform the inverse wavelet transform, we should reformulate it as shown by the schematic diagram in \xf{fig_schematic} (b). In this diagram, the detail subband at each scale is extracted by applying a 1-scale wavelet decomposition on the approximate subband at the upper level (the finer scale). Clearly, this formulation is general, and can be used for arbitrary number of scales. 

\begin{figure}[h!]
	\centering
	\subfloat[][]{
		\includegraphics[width=0.39\textwidth]{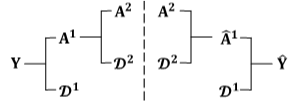}
	}
	
	\subfloat[][]{
		\includegraphics[width=0.44\textwidth]{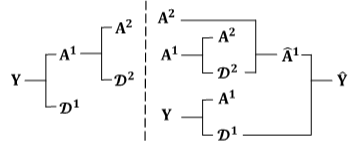}
	}

	\subfloat[][]{
		\includegraphics[width=0.59\textwidth]{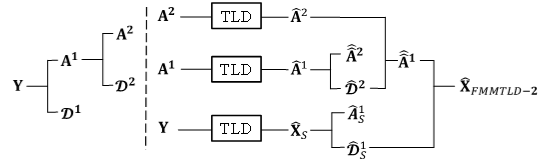}
	}
	\caption{
		Schematic diagrams of the two-scale wavelet transform and FMMTLD-2 methods. (a) Diagram of the common method for computing a two-scale DWT, and its inverse. (b) If we only store the approximate subbands (including the input image itself), 1-scale wavelet transforms are used to extract the detail subbands. (c) Inserting a denoiser (e.g., TLD) at the right side of (b) leads to FMMTLD-2 method which allows for directly denoising the approximate subbands and extracting the detail subbands using 1-scale wavelet transforms. A hat (\^{}) is used above the notation to make symbols with the same names distinguishable.
	}
	\label{fig_schematic}
\end{figure}

The formulation in \xf{fig_schematic} (b) allows us to denoise each detail subband indirectly, and simultaneously perform the subbands mixing. By inserting a denoising method for denoising each approximate subband, the diagram of the resultant method is shown in \xf{fig_schematic} (c). We call it FMMTLD-2. It is easy to see that when the number of scales is set to 1 ($J$ = 1), FMMTLD-2 is exactly equivalent to FMMTLD-1. In \xf{fig:increase} (f) and (h), the outputs of FMMTLD-1 and 2 are visually compared. We list FMMTLD-2 steps in \xalg{alg_fmmtld_2}. Firstly, it stores the noisy image (the finest scale) and all of its approximate subbands in an array (denoted by $\ap$ in \xalg{alg_fmmtld_2}). Secondly, the method starts with the last two approximate subbands ($\ap[s - 1]$ and $\ap[s]$) and denoises them (in lines \ref{alg_fmmtld_2_tld_fine} to \ref{alg_fmmtld_2_tld_coarse}). Thirdly, the method performs the subbands mixing (line \ref{alg_fmmtld_2_S_X}). It reconstructs the denoised approximate subband at the finer scale ($\ap[s - 1]$) by changing its approximate subband with $\ap[s]$ (coarser scale). Finally, these steps are repeated across all scales to reconstruct the final image.

\begin{algorithm}
	\caption{FMMTLD-2}
	\label{alg_fmmtld_2}
	\begin{algorithmic}[1]
		\REQUIRE Noisy image ($\nim$) and its noise level ($\sigma$), Patch size ($p$), Number of scales ($J$).
		\ENSURE Type of wavelet transform ($\wt$) and its filterbank ($af$ and $sf$).
		\STATE $\ap[0] = \{\nim\}$ \hfil \COMMENT{$\ap$ is an array}
		\FOR{s = 1 to $J$}
		\STATE $\sset = \wt(\ap[s-1],1,af)$ \hfil \COMMENT{1-scale $\wt$}
		\STATE $\apsub = $ Get the approximate subband from $\sset$
		\STATE $\ap[s] = \{\apsub\}$
		\ENDFOR
		\FOR{s = $J$ to 1}
		\STATE $\ap[s - 1] = \mathrm{TLD}(\ap[s - 1],\sigma,p)$ \hfil \COMMENT{Fine scale} \label{alg_fmmtld_2_tld_fine}
		\STATE $\ap[s] = \mathrm{TLD}(\ap[s],\sigma,p)$ \hfil \COMMENT{Coarse scale}\label{alg_fmmtld_2_tld_coarse}
		\STATE $\sset_F = \wt(\ap[s - 1], 1 ,af)$
		\STATE $\detset_F = $ Get the detail subbands from $\sset_F$
		\STATE $\sset_C = \wt(\ap[s], 1, af)$
		\STATE $\apsub_C = $ Get the approximate subband from $\sset_C$
		\STATE $S_X = \{ \apsub_C, \detset_F \}$ \label{alg_fmmtld_2_S_X}
		\STATE $\ap[s - 1] = \iwt(\sset_X, 1, sf)$ \label{alg_fmmtld_2_iwt_mix}
		\ENDFOR
		\RETURN $\im_{FMM} = \ap[0]$
	\end{algorithmic}
\end{algorithm}

The same method can be used to fix MMTLD-1 (\xalg{alg_mmtld_1}). In \xalg{alg_mmtld_2}, we list the steps of MMTLD-2 which, in contrast to MMTLD-1, performs the subbands mixing over all scales. Note that when $K$ = 1, this method reduces to MMTLD-1. The outputs of MMTLD-1 and MMTLD-2 are compared in \xf{fig:increase} (e) and (g).

\begin{algorithm}
	\caption{MMTLD-2}
	\label{alg_mmtld_2}
	\begin{algorithmic}[1]
		\REQUIRE Noisy image ($\nim$) and its noise level ($\sigma$), Patch size ($p$), Number of scales ($J$), Number of scales ($K$) for the mixing stage.
		\ENSURE Hyper-parameters of MTLD (see \xalg{alg_mtld}, Type of wavelet transform ($\wt_X$) and its filterbank ($af$ and $sf$).
		\STATE $\im_S = \mathrm{TLD}(\nim,\sigma,p)$
		\STATE $\im_M = \mathrm{MTLD}(\nim,\sigma,p,J)$
		\STATE $\mathrm{temp} = \im_S$
		\FOR{ k  = K to 1}
		\STATE $\hat{\sset}_S = \wt_X(\mathrm{temp},k,af)$
		\STATE $\hat{\detset}_S^k = $ Get the detail subbands from $\hat{\sset}_S$
		\STATE $\hat{\sset}_M = \wt_X(\im_M,k,af)$
		\STATE $\hat{\apsub}_M^k = $ Get the approximate subband from $\hat{\sset}_M $
		\STATE $\hat{S}_X = \{ \hat{\apsub}_M^k, \hat{\detset}_S^k \}$
		\STATE $\mathrm{temp} = \iwt_X(\hat{\sset}_X,k,sf)$
		\ENDFOR
		\RETURN $\im_{MM} = \mathrm{temp}$
	\end{algorithmic}
\end{algorithm}

\subsection{Generalization}\label{methods_gen}

In the previous subsections, we have discussed three multiscale extensions for TLD. These extensions are: MTLD (\xalg{alg_mtld}), MMTLD (\xalg{alg_mmtld_2}), and FMMTLD (\xalg{alg_fmmtld_2}). Although we have introduced our methods in the context of sparsifying transform learning due to its speed and convenient formulation, they are quite general and we can embed another image denoising method in place of TLD. A direct extension can be obtained by replacing TLD with \mbox{K-SVD}. By substituting TLD with \mbox{K-SVD} \cite{Elad_Aharon_2006} and MTLD with \mbox{MS K-SVD} \cite{Ophir_Lustig_Elad_2011} in MMTLD (\xalg{alg_mmtld_2}) and FMMTLD (\xalg{alg_fmmtld_2}), we propose two mixing based multiscale methods for \mbox{K-SVD}. We call them mixed multiscale \mbox{K-SVD} (\mbox{MMK-SVD}) and fast \mbox{MMK-SVD} (\mbox{FMMK-SVD}). 

Although TLD and \mbox{K-SVD} are at the heart of many image denoising methods, these methods and their multiscale extensions are not equipped enough to compete with the state-of-the-art methods. Therefore, we also develop multiscale extensions of a nonlocal method named spatially adaptive iterative singular-value thresholding (SAIST) \cite{Dong_Shi_Li_2013}. SAIST utilizes the nonlocal self-similarity of patches through a simultaneous sparse coding framework \cite{Mairal_Bach_Ponce_Sapiro_Zisserman_2009}. We first develop multiscale SAIST (MSAIST) by replacing TLD with SAIST in \xalg{alg_mtld}. Then, equipped with SAIST and MSAIST, we can directly develop mixed MSAIST (MSAIST) and fast \mbox{MMSAIST} (\mbox{FMMSAIST}) by simply replacing TLD and MTLD in \xalg{alg_fmmtld_2} and \ref{alg_mmtld_2} with SAIST and MSAIST, respectively. In short, we tabulated the baseline denoisers and our proposed multiscale extensions in \xt{tbl_methods}.

\begin{table}[H]
	\caption{
		Baseline denoisers and our proposed multiscale methods
	}
	\label{tbl_methods}
	\begin{center}
		\begin{tabular}{|l|l|}
			\specialrule{0.1em}{0em}{0em}
			Baseline Methods & Our Proposed Mixing Based Multiscale Extensions \\
			\specialrule{0.1em}{0em}{0em}
			TLD \cite{Ravishankar_Bresler_2013a} & MMTLD \\
			MTLD  & FMMTLD \\
			\hline
			\mbox{K-SVD} & \mbox{MMK-SVD} \\
			MS-KSVD \cite{Ophir_Lustig_Elad_2011} & \mbox{FMMK-SVD} \\
			\hline
			SAIST \cite{Dong_Shi_Li_2013} & MMSAIST \\
			MSAIST & \mbox{FMMSAIST} \\
			\specialrule{0.1em}{0em}{0em}
		\end{tabular}%
	\end{center}
\end{table}%

\section{Experimental Results}\label{exp}

In this section, we present the qualitative and quantitative results of the methods introduced in the previous section. The compared methods, datasets, and parameters are thoroughly discussed in the following subsections. The source code to reproduce the experiments will be made publicly available on \href{https://github.com/ashkan-abbasi66}{this website}.

\subsection{Datasets}\label{exp_datasets}

Two datasets are used for removing Gaussian noise from gray-scale images: 1- Classic test images \cite{Papyan_Elad_2016}: includes \textit{Barbara}, \textit{Boat}, \textit{Cameraman}, \textit{Fingerprint}, \textit{Hill}, \textit{Lena}, \textit{Couple}, \textit{Pentagon}, and \textit{Man}. All images have 512x512 pixels, except for \textit{Pentagon} which has 1024x1024 pixels, and 2- CSR test set: a set of relatively high-resolution test images (e.g., 1423x949 pixels) introduced in \cite{Facciolo_Pierazzo_Morel_2017}. We corrupt all of the images from the mentioned datasets by adding white Gaussian noise with three different standard deviations (15, 25, and 50) from relatively weak to strong noise levels. 

\subsection{Compared Methods}\label{exp_methods}
Firstly, the proposed mixing based multiscale methods are compared with their corresponding baselines (\xt{tbl_methods}). In addition, we compare our methods with some of the well-known image denoising methods from the literature. We report the results in two groups. The first group consists of methods which are solely based on local image models. In this group we compare the results of our TLD and \mbox{K-SVD} based multiscale methods with BLS-GSM \cite{Portilla_Strela_Wainwright_Simoncelli_2003}, \mbox{MS K-SVD} \cite{Ophir_Lustig_Elad_2011}, and \mbox{Fused K-SVD} \cite{Sulam_Ophir_Elad_2014}. In the second group, we compare our SAIST based methods with some well-known nonlocal methods including BM3D \cite{Dabov_Foi_Katkovnik_Egiazarian_2007}, SSC-GSM \cite{Dong_Shi_Ma_Li_2015}, STROLLR \cite{strollr_tip2020}, and \mbox{GSRC-NLP} \cite{Zha_Yuan_Wen_Zhou_Zhu_2020}. 

\subsection{Results}\label{exp_results}

\begin{figure*}
	\centering
	\includegraphics[height = 22cm]{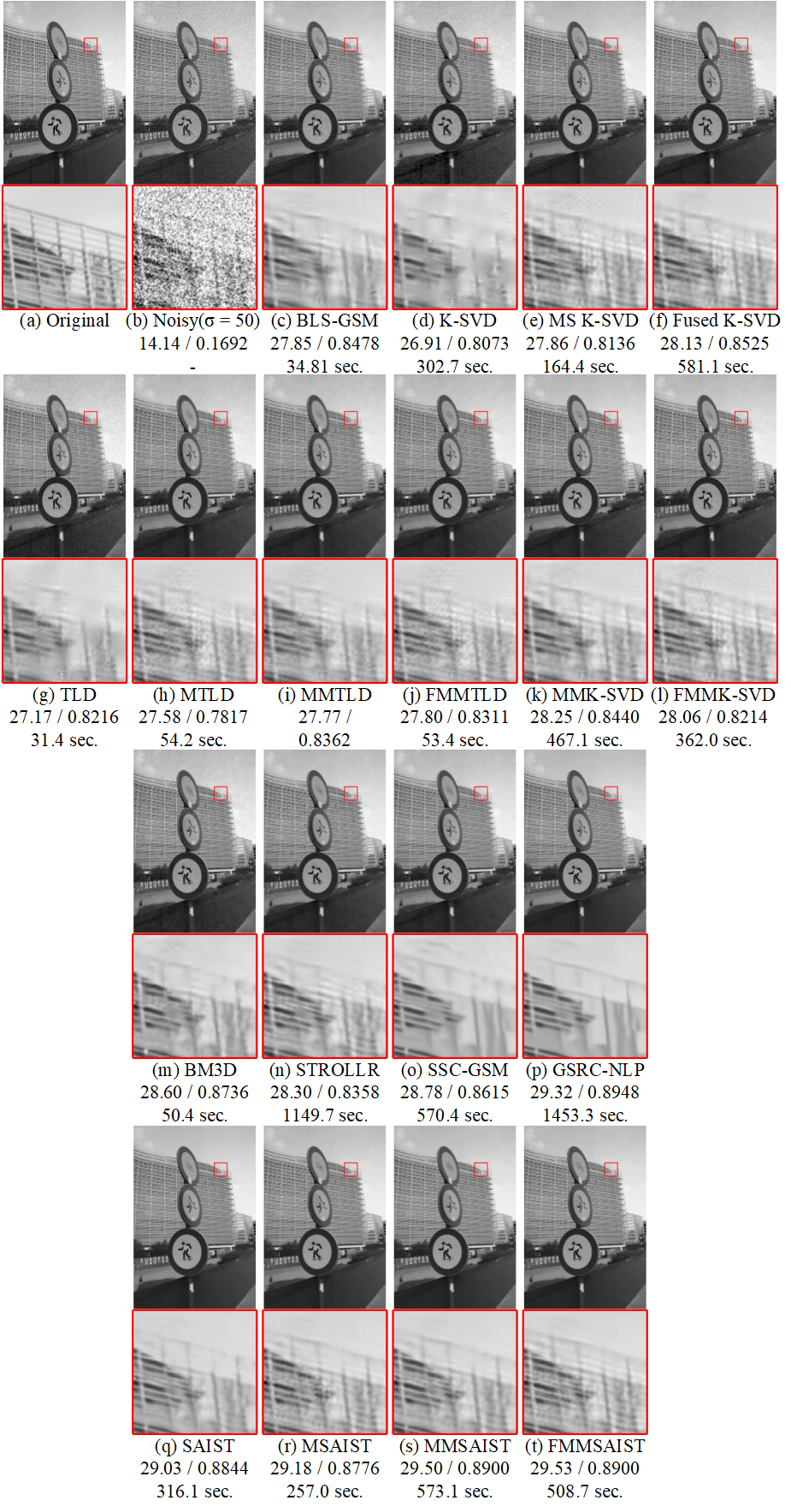}
	\caption{Visual comparison of the results obtained by the compared local ((c) to (l)) and nonlocal ((m) to (t)) image denoising methods. For each method, its PSNR, SSIM, and runtime are reported below it.}
	\label{fig_5210}
\end{figure*}

\xf{fig_lena} and \ref{fig_5210} show some visual comparisons between the denoising outputs of the compared methods. The input image in \xf{fig_lena} (a) is corrupted with a medium noise level ($\sigma$ = 25). The results of local methods and nonlocal methods are denoted by ((c) to (l)) and ((m) to (t)), respectively. It can be seen that both MMTLD (i) and FMMTLD (j) improve the quality of denoised images significantly in comparison with their baselines (TLD (g) and MTLD (h)). The artifacts are responsible for the poor performance of MTLD and this weakness is mitigated in our proposed mixing based methods. However, both BLS-GSM (c) and \mbox{Fused K-SVD} (f) perform better than our best TLD based method (MMTLD). This is because of the poor performances of the baselines (TLD and MTLD) on this image. For this image, \mbox{K-SVD} (d) and \mbox{MS K-SVD} (e) are better than TLD and MTLD. If we replace TLD and MTLD with \mbox{K-SVD} and \mbox{MS K-SVD}, the resultant methods (\mbox{MMK-SVD} (k) and \mbox{FMMK-SVD} (l)) have competitive performances in comparison with BLS-GSM and \mbox{Fused K-SVD}. Note that the runtimes of our \mbox{MMK-SVD} and \mbox{FMMK-SVD} are significantly lower than \mbox{Fused K-SVD} while their performances are very competitive. In the second group (the nonlocal methods), single-scale SAIST (q) cannot compete with BM3D (m), STROLLR (n), SSC-GSM (o), and \mbox{GSRC-NLP} (p). However, our proposed mixing based multiscale SAIST methods are able to achieve comparable performances within reasonable runtimes.

\begin{figure*}
	\centering
	\includegraphics[height = 22cm]{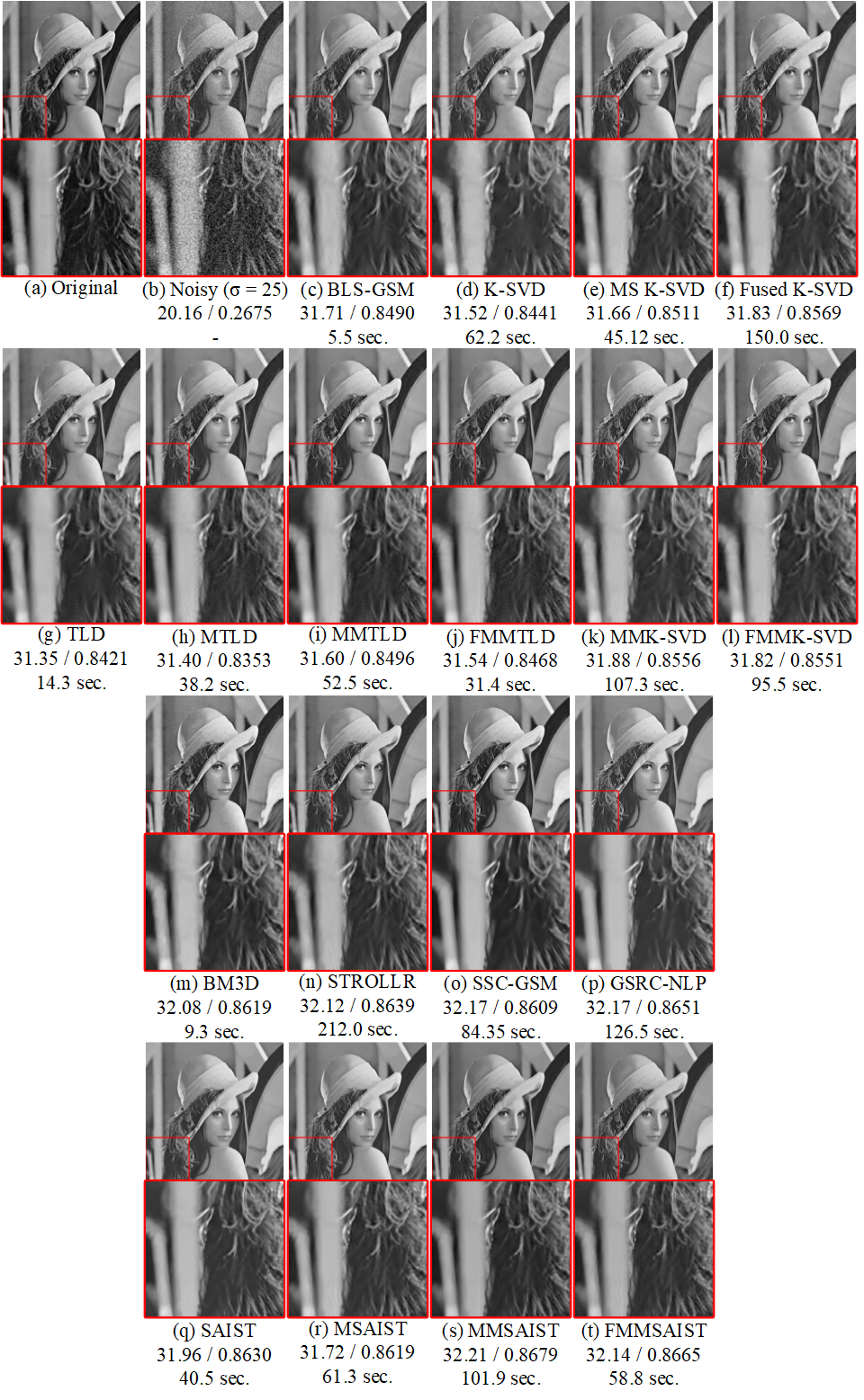}
	\caption{Visual comparison of the results obtained by the compared local ((c) to (l)) and nonlocal ((m) to (t)) image denoising methods. For each method, its PSNR, SSIM, and runtime are reported below it.}
	\label{fig_lena}
\end{figure*}

The input image in \xf{fig_5210} (b) is corrupted with a relatively strong noise ($\sigma$=50). Since this image has a high spatial resolution, the runtimes of the compared methods grow substantially. In the first group (local methods), the three best denoising results, in terms of PSNR, are obtained in order by our \mbox{MMK-SVD} (k), \mbox{Fused K-SVD} (f), and our \mbox{FMMK-SVD} (l). Note that our results are more than 100 seconds faster than \mbox{Fused K-SVD}, while they are very comparable to it. In the nonlocal methods, clearly our methods (MMSAIST (s) and \mbox{FMMSAIST} (t)) outperform the others within reasonable runtimes. For this image, the PSNR improvements of our MMSAIST and \mbox{FMMSAIST} over their single-scale baseline (SAIST (q)) are about 0.5 decibels.

\begin{table}[h!]
	\centering
	\caption{
		Mean of PSNR (dB) and runtime of each method for Gaussian image denoising of classic test images. For each group of compared methods, the best PSNR result is shown in \textbf{bold} and the second one is \underline{underlined}.
	}
	\setlength\tabcolsep{2pt}
	\begin{tabular}{lcccc}
		\toprule
		Methods      & 15                & 25                & 50                & Runtime \\
		\midrule
		Noisy Images & 24.61             & 20.17             & 14.15             & -       \\
		BLS-GSM      & 31.96             & 29.56             & 26.53             & 7.6     \\
		TLD          & 32.03             & 29.45             & 26.10             & 18.8    \\%
		MTLD         & 31.87             & 29.46             & 26.30             & 52.7    \\
		MMTLD        & 32.11             & 29.64             & 26.47             & 71.5    \\
		FMMTLD       & 32.05             & 29.59             & 26.43             & 43.6    \\
		K-SVD        & 32.13             & 29.58             & 26.00             & 87.0    \\
		\mbox{MS K-SVD}     & 31.95             & 29.65             & 26.68             & 81.4    \\
		\mbox{Fused K-SVD}  & \underline{32.20} & \underline{29.84} & \textbf{26.78}    & 265.3   \\%
		MMK-SVD     & \textbf{32.23}    & \textbf{29.90}    & 26.68             & 169.3   \\
		FMMK-SVD    & 32.11             & 29.81             & \underline{26.69} & 136.2   \\
		\hline
		BM3D  \rule{0cm}{2.5 ex}
		& 32.47             & 30.16             & 27.05             & 12.5      \\
		STROLLR      & 32.38             & 30.05             & 26.98             & 286.1     \\
		SSC-GSM      & \underline{32.52} & 30.04             & 27.06             & 110.5     \\
		\mbox{GSRC-NLP}     & \textbf{32.53}    & \textbf{30.20}    & 27.15             & 219.8     \\
		SAIST        & 32.16             & 30.03             & 27.03             & 60.3      \\
		MSAIST       & 31.68             & 29.60             & 26.98             & 76.0      \\
		MMSAIST      & 32.34             & \textbf{30.20}    & \textbf{27.30}    & 136.3     \\
		\mbox{FMMSAIST}     & 32.30             & \underline{30.16} & \underline{27.20} & 93.2      \\
		\bottomrule
	\end{tabular}
	\label{tbl_classic}
\end{table}

The observations and explanations can be validated by the average quantitative results. The average of PNSR results obtained by the compared methods on the classic test images are reported in \xt{tbl_classic}. It can be seen that in almost all cases our proposed mixing based multiscale methods improve significantly over their baselines. In the first group (local methods), our \mbox{MMK-SVD} method generally outperforms the others. In terms of average runtime, TLD based methods are significantly faster than \mbox{K-SVD} based methods, however, their results are inferior to them. In the second group, although our baseline (SAIST) is inferior to all other compared methods (BM3D, SSC-GSM, STROLLR and GSRC-NLP), our proposed mixing based multiscale SAIST methods can favorably compete with them. When noise level is weak, SSC-GSM and \mbox{GSRC-NLP} achieve the best results, and STROLLR, MMSAIST, and \mbox{FMMSAIST} are very close to each other. When noise is not weak, our MMSAIST and \mbox{FMMSAIST} methods perform better than the others within reasonable runtimes. 

In \xt{tbl_csr}, we report the average PSNR and runtime for all of the compared methods on the task of removing Gaussian noise from CSR test set \cite{Facciolo_Pierazzo_Morel_2017}. In the first group, the best performances are achieved by \mbox{MMK-SVD} and \mbox{Fused K-SVD}. Although their performances are nearly the same, \mbox{MMK-SVD} is more than 250 seconds faster than \mbox{Fused K-SVD}. In the second group, MMSAIST and \mbox{FMMSAIST} outperform the other methods when noise is not weak. Interestingly, it can be seen that the average runtime of MMSAIST and \mbox{FMMSAIST} are significantly lower than their corresponding \mbox{K-SVD} based methods (\mbox{MMK-SVD} and \mbox{FMMK-SVD}) while SAIST methods can provide competitive results in comparison with the state-of-the-art methods such as SSC-GSM and \mbox{GSRC-NLP}. 

\begin{table}[h!]
	\centering
	\caption{
		Mean of PSNR (dB) and runtime of each method for Gaussian image denoising of gray-scale high-resolution images of CSR test set. For each group of compared methods, the best PSNR result is shown in \textbf{bold} and the second one is \underline{underlined}.
	}
	\setlength\tabcolsep{2pt}
	\begin{tabular}{lcccc}
		\toprule
		Methods      & 15                & 25                 & 50               & Runtime   \\
		\midrule
		Noisy Images & 24.61             & 20.17              & 14.16            & -         \\
		BLS-GSM      & 33.59             & 30.91              & 27.63            & 30.9      \\
		TLD          & 33.36             & 30.49              & 26.90            & 36.4      \\
		MTLD         & 33.31             & 30.61              & 27.15            & 65.9      \\
		MMTLD        & 33.53             & 30.78              & 27.37            & 102.3     \\
		FMMTLD       & 33.47             & 30.74              & 27.33            & 68.0      \\
		K-SVD       & 33.59             & 30.57              & 26.83            & 311.6     \\
		\mbox{MS K-SVD}     & 33.27             & 30.63              & 27.41            & 278.3     \\
		\mbox{Fused K-SVD}  & \underline{33.71} & \underline{30.97}  & \underline{27.64}& 848.5     \\
		MMK-SVD     & \textbf{33.74}    & \textbf{31.06}     & \textbf{27.67}   & 590.0     \\
		FMMK-SVD    & 33.48             & 30.81              & 27.47            & 438.9     \\
		\hline
		BM3D  \rule{0cm}{2.5 ex}       
		& 34.07             & 31.42              & 27.98             & 47.0      \\
		STROLLR      & 33.88             & 31.22              & 27.82             & 1096.8    \\
		SSC-GSM      & \underline{34.13} & 31.30              & 27.91             & 407.6     \\
		\mbox{GSRC-NLP}     & \textbf{34.14}    & \underline{31.48}  & 28.04             & 801.7     \\
		SAIST        & 33.40             & 31.28              & 27.89             & 227.2     \\
		MSAIST       & 33.38             & 31.10              & 28.09             & 257.4     \\
		MMSAIST      & 33.83             & \textbf{31.54}     & \textbf{28.26}    & 484.6     \\
		\mbox{FMMSAIST}     & 33.80             & \underline{31.48}  & \underline{28.20} & 338.7     \\
		\bottomrule
	\end{tabular}
	\label{tbl_csr}
\end{table}

\subsection{Overall Gain Comparisons}\label{exp-gauss-gain}
The entries of \xt{tbl_classic} and \ref{tbl_csr} show that the performance gains obtained by our proposed mixing based multiscale methods over their corresponding baselines not only depend on the noise level, but also on the baseline methods and datasets. On the classic test images, when noise is relatively weak ($\sigma = 15$), FMMTLD and \mbox{FMMK-SVD} are slightly inferior than their corresponding baselines (TLD and \mbox{K-SVD}, respectively). However, for all noise levels, MMTLD and \mbox{MMK-SVD} are better than their baselines. This is also true for CSR test set. Comparing gains of \mbox{MMK-SVD} with \mbox{Fused K-SVD} over pure \mbox{K-SVD} on both datasets (\xt{tbl_classic} and \ref{tbl_csr}) show that in the most of cases \mbox{MMK-SVD} acheives bigger gain while it is considerably faster than \mbox{Fused K-SVD}. In addition, comparing the gains obtained by \mbox{Fused K-SVD} and our proposed mixing based multiscale methods over their corresponding baselines on both datasets show that bigger gains are obtained on the CSR test images. This shows that images with bigger spatial size are more likely to benefit from multiscale processing.

Using \xt{tbl_classic} and \ref{tbl_csr}, we plot the average PSNR gains for SAIST based methods in \xf{fig_plot_SAIST_gains}. It is evident that both \mbox{MMSAIST} and \mbox{FMMSAIST} favorably outperform pure SAIST for all noise levels on both datasets. Note that even when there is a big performance gap between SAIST and MSAIST (e.g., $\sigma = 25$ in \xf{fig_plot_SAIST_gains} (a)), our mixing based methods are able to combine the benefits of both methods and achieve significant improvement over them. 
\begin{figure}[t!]
	\centering
	\subfloat[The gains obtained on the classic test images]{\includegraphics[width=0.4\textwidth]{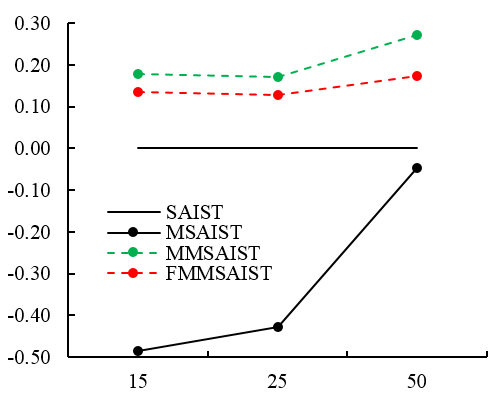}}
	\hfil
	\subfloat[The gains obtained on the CSR test set]{\includegraphics[width=0.4\textwidth]{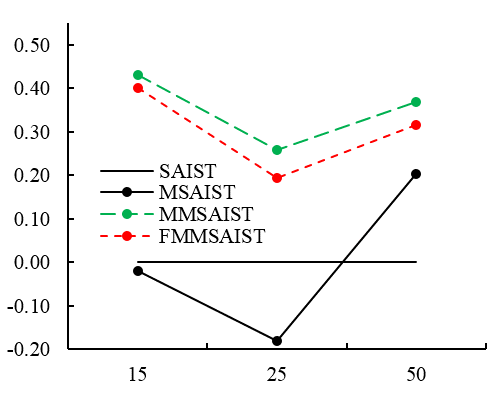}}	
	\caption{The averaged PSNR gains obtained by the multiscale SAIST based methods over the pure SAIST on the (a) classic test images, and (b) CSR test set. For each plot, the horizontal axis shows different noise levels, and the vertical axis shows the average PSNR gains. The colored dashed lines represent the gains obtained by the mixing based multiscale methods (MMSAIST and \mbox{FMMSAIST}).}
	\label{fig_plot_SAIST_gains}
\end{figure}

\subsection{Effects of Number of Decomposition Levels}\label{exp_levels}
The criterion used in \mbox{MS K-SVD} \cite{Ophir_Lustig_Elad_2011} and \mbox{Fused K-SVD} \cite{Sulam_Ophir_Elad_2014} for selecting the number of scales solely relied on the spatial size of the images. They experimentally observed that using these methods with more than 1 scale negatively affects the quality of denoising for images with width or height less than or equal to 550 pixels. Similar observations were also made in \cite{Hughes_Rockmore_Wang_2013, Mairal_Sapiro_Elad_2008, Papyan_Elad_2016} where the authors have reported their final results based on using decompositions with 1 or 2 scales. 

In this section, we want to explicitly show how the performances are changed as the number of decomposition levels increases. To this end, let\textquotesingle s denoise the \textit{Pentagon} test image corrupted by a Gaussian noise with $\sigma = 50$. For this image, \mbox{MS K-SVD} and \mbox{Fused K-SVD} use wavelet transforms with 2 scales. Let\textquotesingle s denoise this image with TLD, \mbox{K-SVD}, and their multiscale counterparts. Keeping all of the parameters fixed except for the number of decomposition levels, we plot the average PSNR gains of the multiscale methods over their corresponding single-scale baseline in \xf{fig_plot_num_levels}.

\begin{figure}[ht!]
	\centering
	\subfloat[]{\includegraphics[width=0.4\textwidth]{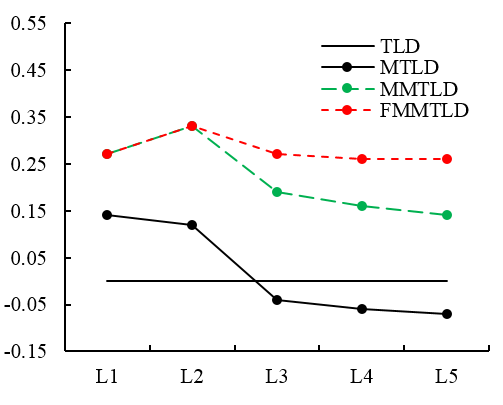}}
	\hfil
	\subfloat[]{\includegraphics[width=0.4\textwidth]{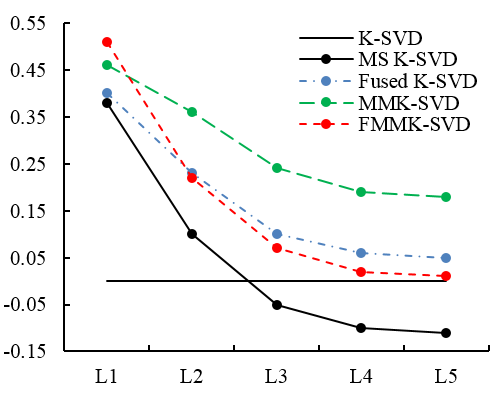}}	
	\caption{The PSNR gains obtained by varying the number of decomposition levels for the TLD and \mbox{K-SVD} based multiscale methods over their single-scale counterparts. The test image is \textit{Pentagon} corrupted by Gaussian noise with $\sigma = 50$. In (a), the gains obtained by the multiscale TLD based methods are reported. In (b), the gains obtained by the multiscale \mbox{K-SVD} based methods are reported. For each plot, the horizontal axis shows the number of decomposition levels, and the vertical axis shows the PSNR gains. The colored dashed lines represent the gains obtained by the fusion/mixing based multiscale methods.}
	\label{fig_plot_num_levels}
\end{figure}

These plots (\xf{fig_plot_num_levels}) show that increasing the number of scales to more than one is just slightly beneficial for the noisy \textit{Pentagon} when it is denoised through MMTLD or FMMTLD. In \xf{fig_plot_num_levels} (b), it can be seen that the efficiencies of all \mbox{K-SVD} based methods drop when they are used with more than one scale. Note that our proposed \mbox{MMK-SVD} method behaves better than \mbox{Fused K-SVD} while it is faster than \mbox{Fused K-SVD}. Also, when one scale is used \mbox{FMMK-SVD} achieves more gain than \mbox{MMK-SVD} and \mbox{Fused K-SVD}. In short, similar to \cite{Ophir_Lustig_Elad_2011,Sulam_Ophir_Elad_2014,Hughes_Rockmore_Wang_2013,Mairal_Sapiro_Elad_2008,Papyan_Elad_2016}, our experiments also suggest that using only one or two scales is sufficient.

\subsection{Effects of Pattern Sizes: Denoising a Checkerboard Image}\label{exp_pattern_sizes}
It is reasonable that the image content is an important factor to determine whether multiscale denoising is beneficial or not. Here, we want to explicitly study pattern sizes and the importance of multiscale processing. Therefore, we create a synthetic checkerboard image (shown in \xf{fig_checkerboard} (a)) which has five regions. Each region contains black and white square patterns with the same size. Next, we corrupt it by Gaussian noises with $\sigma=25$ and $ 50$. Then, we use our TLD based methods to explicitly show that where multiscale patch-based denoising is beneficial. The PSNR gains of the multiscale methods over the pure TLD are plotted in \xf{fig_checkerboard} (b) and (c). Note that the actual patch size of TLD is $11\times 11$ pixels. 

\begin{figure}[ht!]
	\centering
	\subfloat[]{\includegraphics[width=0.8\textwidth]{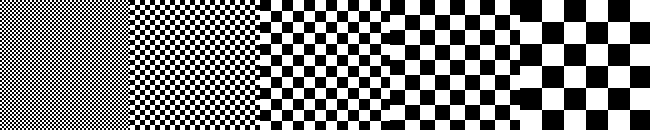}}
	
	\subfloat[]{\includegraphics[width=0.4\textwidth]{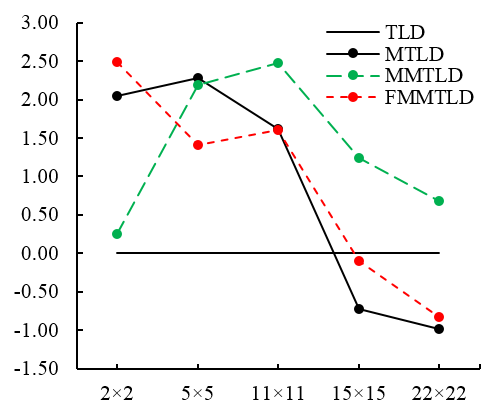}}	
	\hfil
	\subfloat[]{\includegraphics[width=0.4\textwidth]{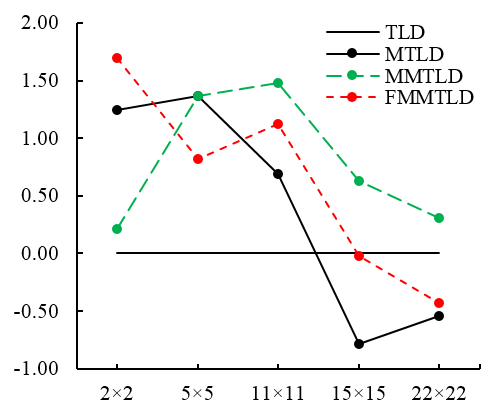}}	
	\caption{The plots of PSNR gains obtained by the multiscale methods over their single-scale counterparts for denoising the checkerboard image (a) corrupted by two Gaussian noises. In (b), the image is corrupted by a noise with $\sigma = 25$, and in (c), the image is corrupted by a noise with $\sigma = 50$. For each plot, the vertical axis shows PSNR gains and the horizontal axis shows sizes of the square patterns in the checkerboard image which divides it into five regions.}
	\label{fig_checkerboard}	
\end{figure}

These plots show that wherever the square patterns are smaller than or equal to the patch size ($11\times 11$), the multiscale denoising is more beneficial. This is reasonable since when the patterns are big, learning a sparsifying transform to reconstruct big homogeneous regions becomes an easy task. In this case, further processing by multiscale methods negatively affects the results. On the one hand, the effectiveness of MTLD and FMMTLD in recovering small details are comparable or even better than MMTLD. On the other hand, MMTLD has positive, albeit small, contributions for recovering all patterns. This partially shows why MMTLD has a better performance in our previous experiments.

\subsection{Runtimes}\label{exp_runtimes}
Throughout this paper, we reported the runtimes of the compared methods in almost all experiments. The methods were implemented in MATLAB, and the experiments were conducted on a desktop PC with an Intel® i5-7400K CPU at 3.0 GHz and 16 GB of RAM.

Each multiscale method in \xt{tbl_methods} relies on a single-scale baseline for which the computational complexity is discussed thoroughly in its original paper. Therefore, here, we only focus on the multiscale methods irrespective of their corresponding single-scale baseline. However, to make the statements more concrete, let\textquotesingle s consider the multiscale methods based on TLD. Then, the key computations of MTLD (\xalg{alg_mtld}), MMTLD (\xalg{alg_mmtld_2}), and FMMTLD (\xalg{alg_fmmtld_2}) are as follows: 
\begin{inparaenum}
	\item
	forward wavelet transform, 
	\item
	inverse wavelet transform, and 
	\item
	denoising through TLD. 
\end{inparaenum}
From these operations, it is evident that the most time-consuming one is TLD. The computational complexity of TLD is addressed in its original papers (e.g., \cite{Ravishankar_Bresler_2013a}). 

In MTLD, if we use a \jscale wavelet transform, the method runs TLD to denoise $3J+1$ subbands (including the last approximate subband). To implement MTLD, we use DWT which is a non-redundant wavelet transform. Therefore, the wavelet subbands are smaller than the input image, and it is expected that the multiscale denoising via MTLD does not significantly increase the overall computations. However, in practice, the runtime of MTLD is higher than TLD (e.g., see their corresponding runtimes in \xt{tbl_classic}) since denoising subbands takes more time than we expected. This is because TLD\textquotesingle s runtime not only depends on the size of its input image but also on its content and noise level \cite{Ravishankar_Bresler_2015}. 

The runtime of MMTLD is dominated by the runtimes of TLD and MTLD since MMTLD\textquotesingle s mixing stage is just involved with some cheap operations (forward wavelet transform and its inverse). In \xt{tbl_classic} and \ref{tbl_csr}, it can be seen that by adding the runtimes of TLD and MTLD, the runtime of MMTLD can be well approximated. This shows that the whole mixing stage takes less than 0.1 second for the two datasets reported in this table. 

The main advantage of FMMTLD over MMTLD comes from the fact that it saves operations needed to denoise the detail subbands. Considering the fact that bottleneck of MTLD and MMTLD are denoising wavelet subbands, it would be clear that FMMTLD suggests a significant reduction in the computational complexity which comes at the cost of restricting both of the denoising and mixing stages to use the same type of wavelet transform. Since FMMTLD works by denoising the low-pass subbands (including the input image itself), it is theoretically 1 + 1/4 + 1/16 + … = 4/3 more complex than TLD. However, due to the convergence issues of TLD that we have mentioned earlier for MTLD, FMMTLD\textquotesingle s runtime is slightly bigger in practice.

\section{Conclusion}\label{conc}
The data-driven sparse models (such as synthesis dictionary learning \cite{Aharon_Elad_Bruckstein_2006,Elad_Aharon_2006} and sparsifying transform learning \cite{Ravishankar_Bresler_2013a,Ravishankar_Bresler_2013b, Ravishankar_Bresler_2015} are used at the heart of many image denoising methods. However, these models are intrinsically single-scale. In this paper, we investigate several methods to integrate multiscale analysis with the data-driven sparse models. To this end, we propose two methods developed based on wavelet subbands mixing. In contrast to \cite{Ophir_Lustig_Elad_2011,Sulam_Ophir_Elad_2014,Yan_Shao_Liu_2013}, we show that without directly denoising detail subbands, the multiscale processing using wavelet transform is achievable, and we can efficiently combine the advantages of both single-scale and multiscale methods while avoiding their weaknesses. Our experiments show that the proposed mixing based methods are capable of recovering the underlying image details and greatly suppress artifacts. Although we establish our methods in the context of sparsifying transform learning due to its convenient formulation, practical efficiency, and numerical stability; we generalize the studied methods to propose our mixing based multiscale methods for \mbox{K-SVD} denoising \cite{Elad_Aharon_2006} and SAIST \cite{Dong_Shi_Li_2013}. On the task of Gaussian image denoising, we show that although plain SAIST cannot compete with the state-of-the-art methods such as SSC-GSM \cite{Dong_Shi_Ma_Li_2015} and \mbox{GSRC-NLP} \cite{Zha_Yuan_Wen_Zhou_Zhu_2020}, our proposed multiscale extensions for SAIST can achieve competitive results within reasonable runtimes. We highlight the limitations of the proposed methods through extensive experiments. They can serve as promising research directions. A future extension to this paper might be using some modern denoising approaches such as a deep learning based denoising method \cite{Zhang_Zuo_Chen_Meng_Zhang_2017}. Moreover, because removing the need for denoising detail subbands can reduce runtime and computational complexity, the proposed approach has a potential to be extended for three dimensional denoising. Also, extending these methods for other inverse problems (e.g., super-resolution) can also be considered as a research direction. 


%
%
%
%


\section*{Acknowledgment}
The authors also would like to thank Prof. Ravishankar and his colleagues for making their source code freely available.

\bibliographystyle{ieeetr}
\bibliography{mtld_bibs}  






\end{document}